\documentclass[11pt, oneside]{article}    

\usepackage[final]{nips_2017}

\usepackage{geometry}
\usepackage[utf8]{inputenc} 
\usepackage[T1]{fontenc}    
\usepackage{hyperref}       
\usepackage{url}            
\usepackage{booktabs}       
\usepackage{amsfonts}       
\usepackage{nicefrac}       
\usepackage{microtype}      
\usepackage{tabularx}
\usepackage{caption}
\usepackage{mdwlist}
\usepackage{amssymb}
\usepackage{diagbox}
\usepackage{subfig}
\usepackage{natbib}
\usepackage{amsmath,amssymb,graphicx}

\title{Learning to Run with Actor-Critic Ensemble}
\date{\today}

\author{Zhewei Huang$^{1,2}$, Shuchang Zhou$^{1}$, BoEr Zhuang$^{1,2}$, Xinyu Zhou$^{1}$\\
1. Megvii Inc.\\
2. Peking University\\
\texttt{\{huangzhewei,zsc,zhuangboer,zxy\}@megvii.com} \\
}

\usepackage{cite}
\begin{document}
\maketitle
\section{Method}
We introduce an Actor-Critic Ensemble(ACE) method for improving the performance of Deep Deterministic Policy Gradient(DDPG) algorithm\citep{lillicrap2015continuous}. At inference time, our method uses a critic ensemble to select the best action from proposals of multiple actors running in parallel. By having a larger candidate set, our method can avoid actions that have fatal consequences, while staying deterministic. Using ACE, we have won the 2nd place in NIPS'17 Learning to Run competition, under the name of "Megvii-hzwer"\footnote{\url{https://www.crowdai.org/challenges/nips-2017-learning-to-run/leaderboards}} .

\subsection{ACE}
\subsubsection{Dooming Actions Problem of DDPG}
The competition of Learning to Run asks the participants to teach a skeleton with legs to run as far as possible in 1000 steps, while avoiding random obstacles on the ground. In the game, we find that legs of a fast running skeleton can easily be tripped by obstacles, which causes the skeleton to enter an unstable state with limbs swing widely and fall down after a few frames. We call the action causing the skeleton to enter unstable state a "dooming action", as it is almost impossible to recover from the unstable states.

\begin{figure*}[ht!]%
        \centering
        \subfloat[DDPG and ACE]{
                \includegraphics[width=0.5\textwidth]{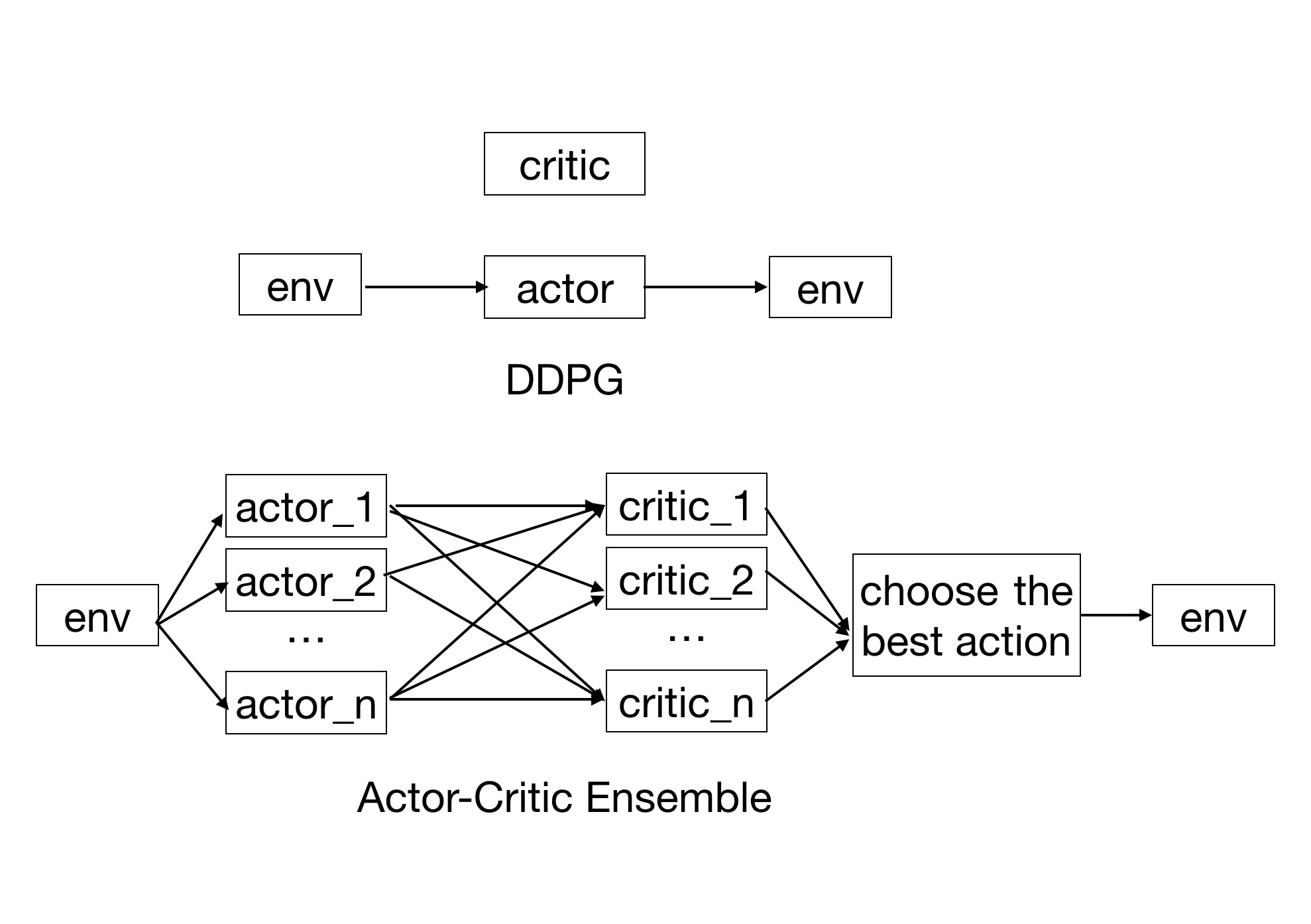}
                \label{fig:ddpg-ace}
        }
        \subfloat[Performance of ACE]{
                \includegraphics[width=0.5\textwidth]{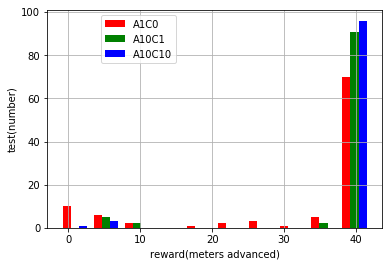}
                \label{fig:performance}
        }
        \caption{Schema for DDPG and ACE}
        \label{fig:ace}        
\end{figure*}

To investigate dooming actions, we let the critic network inspect the actions at inference time. We find that most of the time, the critic can recognize dooming actions by giving low scores. However, as there is only one action proposed by the actor network in DDPG at every step, the dooming actions cannot be avoided. This observation leads us to propose using an actor ensemble to allow the agent to avoid dooming actions by having a critic ensemble to pick the best action, as shown in Fig.~\ref{fig:ace}(a).

\subsubsection{Inference-Time Actor Critic Ensemble}
We first train multiple actor-critic pairs separately, using the standard DDPG method. Then we build a new agent with many actor networks proposing actions at every step. Given multiple actions, a critic network is used to select the best action. We simply pick the action with the highest score, and send it to the actuator.

Empirically, we find that using actors of heterogeneous nature, like trained with different hyper-parameters, to perform better than using actors from different epochs of the same training setting. This is in agreement with the observations in Ensemble Learning\citep{dietterich2000ensemble}.

To further improve the prediction quality of the critic, we build an ensemble of critics, by picking the pairing critics of actors. The outputs of the critic networks are combined by taking average.

\subsubsection{Training with ACE}
If we put actor networks together to train, all the actor networks can be updated at every step, even if its action is not used. The modified Bellman equation is as below:
\begin{center}
$i_{t+1} = \arg\max_j{Q(s_{t+1},\mu_j(s_{t+1}))}$ \\
$Q(s_t, a_t) = r(s_t, a_t) + \gamma Q(s_{t+1},\mu_{i_{t+1}}({s_{t+1}}))$.
\end{center}

\section{Experiments}
\subsection{Baseline Implementation}
We use the DDPG as our baseline. To describe the state of the agent, we collect three consecutive frames of observations from the environment. These information goes through the feature engineering as proposed by Yongliang Qin\footnote{\url{https://github.com/ctmakro/stanford-osrl}} before being fed into the network.

\begin{figure*}[ht!]%
        \centering
        \subfloat[Reward per episode(60 processes)]{
                \includegraphics[width=0.5\textwidth]{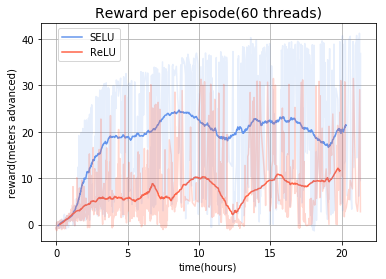}
                \label{fig:relu-selu-60}
        }
        \subfloat[Reward per episode(20 processes)]{
                \includegraphics[width=0.5\textwidth]{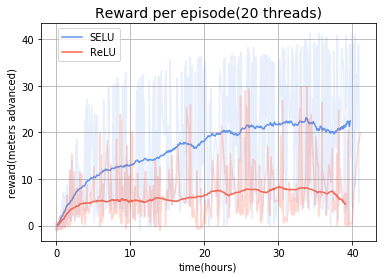}
                \label{fig:relu-selu-20}
        }
        \caption{Training with different activation functions and different number of processes for generating training data, by DDPG}
        \label{fig:ablation}
\end{figure*}

As the agent is expected to run 1000 steps to finish a successful episode, we find the vanishing gradient problem to be critical. We make several attempts to deal with this difficulty. First, we find that with the original simulation timestep, the DDPG converges slowly. In contrast, using four times larger simulation timestep, which is equivalent to changing the action only every four frames, is found to speedup convergence significantly. We have also tried unrolling DDPG as in $TD(\lambda)$ with $\lambda=4$\citep{anonymous2018distributional}, but found it be inferior to simply increasing simulation timestep. Second, we have tried several activation functions and found that the activation function of Scaled Exponential Linear Units(SELU)\citep{klambauer2017self}, to be superior to ReLU, Leaky ReLU, Tanh and Sigmoid, as shown in Fig.~\ref{fig:ablation}.

\begin{table}
\caption{Hyper-parameters used in the experiments}
\begin{tabularx}{\columnwidth}{r|X}
  \toprule
  Actor network architecture & $[FC800, FC400]$, Tanh for output layer and SELU for other layers\\
  Critic network architecture & $[FC800, FC400]$, linear for output layer and SELU for other layers\\
  Actor learning rate & 3e-4 \\
  Critic learning rate & 3e-4 \\
  Batch size & 128 \\
  $\gamma$ & 0.96 \\
  replay buffer size & $2\mathrm{e}{6}$\\
                          \bottomrule
\end{tabularx}
\label{tab:hyperopt}
\end{table}

\subsection{ACE experiments}
For all models we use an identical architecture of actor and critic networks, with hyper-parameters listed in Table~\ref{tab:hyperopt}. Our code used for competition can be found online\footnote{\url{https://github.com/hzwer/NIPS2017-LearningToRun}}.
  
We build the ensemble by drawing models trained with settings of the last section. Fig.~\ref{fig:ace}(b) gives the distribution of rewards when using ACE, where AXCY stands for X number of actors and Y number of critics. It can be seen that A10C10 (having 10 critics and 10 actors) has a much smaller chance of falling (rewards below 30) compared to A1C0, which is equivalent to DDPG. The maximum rewards also get improved, as shown in Tab.~\ref{tab:ace}.

Training with ACE is found to have similar performance as Inference-Time ACE.

\begin{table}
\centering
\caption{Performance of ACE}
\label{tab:ace}
\begin{tabular}{@{}ccccclc@{}}
\toprule
Experiment & \# Test & \# Actor & \# Critic & Average reward & Max reward & \# Fall off \\ \midrule
A1C0 & 100 & 1 & 0 & 32.0789 & 41.4203 & 25 \\
A10C1 & 100 & 10 & 1 & 37.7578 & 41.4445 & 7 \\
A10C10 & 100 & 10 & 10 & 39.2579 & 41.9507 & 4 \\ \bottomrule
\end{tabular}
\end{table}

\section{Conclusion}
We propose Actor-Critic Ensemble, a deterministic method that avoids dooming actions at inference time by asking an ensemble of critics to pick actions proposed by an ensemble of actors. Experiments find that ACE can significantly improve the performance of DDPG, exhibited by less number of fallings and increased speed of the running skeletons.

\nocite{*}
\bibliographystyle{plainnat}
\bibliography{writeup}
\end{document}